\documentclass{article}

\usepackage[nonatbib,final]{neurips_2022}

\usepackage[utf8]{inputenc} 
\usepackage[T1]{fontenc}    
\usepackage{nicefrac}
\usepackage{wrapfig}
\usepackage{chngcntr}
\usepackage{color,xcolor}
\usepackage{epsfig}
\usepackage{graphicx}

\usepackage{adjustbox}
\usepackage{array}
\usepackage{booktabs}
\usepackage{colortbl}
\usepackage{float,wrapfig}
\usepackage{hhline}
\usepackage{multirow}
\usepackage{subcaption} 

\usepackage{amsmath,amsfonts,amsthm,amssymb}
\usepackage{bm}
\usepackage{nicefrac}
\usepackage{microtype}

\usepackage{changepage}
\usepackage{extramarks}
\usepackage{fancyhdr}
\usepackage{lastpage}
\usepackage{setspace}
\usepackage{soul}
\usepackage{xspace}

\usepackage{hyperref}
\hypersetup{colorlinks, breaklinks, citecolor=[rgb]{0.6, 0.6, 1.0},
anchorcolor=[rgb]{0.6, 0.6, 1.0}, linkcolor=[rgb]{0.6, 0.6, 1.0},
}
\usepackage[nocompress]{cite}
\usepackage{url}

\usepackage{enumerate}

\usepackage{titlesec}
\usepackage{listings}

\usepackage{enumitem}

\newcolumntype{L}[1]{>{\raggedright\let\newline\\\arraybackslash\hspace{0pt}}m{#1}}
\newcolumntype{C}[1]{>{\centering\let\newline\\\arraybackslash\hspace{0pt}}m{#1}}
\newcolumntype{R}[1]{>{\raggedleft\let\newline\\\arraybackslash\hspace{0pt}}m{#1}}

\newcommand{\ignorethis}[1]{}

\makeatletter
\DeclareRobustCommand\onedot{\futurelet\@let@token\@onedot}
\def\@onedot{\ifx\@let@token.\else.\null\fi\xspace}

\def\eg{\emph{e.g}\onedot} 
\def\ie{\emph{i.e}\onedot} 
 
\def\etc{\emph{etc}\onedot} 
\def\wrt{w.r.t\onedot} 

\makeatother

\definecolor{mydarkblue}{rgb}{0,0.08,1}
\definecolor{mydarkgreen}{rgb}{0.02,0.6,0.02}
\definecolor{mydarkred}{rgb}{0.8,0.02,0.02}
\definecolor{mydarkorange}{rgb}{0.40,0.2,0.02}
\definecolor{mypurple}{RGB}{111,0,255}
\definecolor{myred}{rgb}{1.0,0.0,0.0}
\definecolor{mygold}{rgb}{0.75,0.6,0.12}
\definecolor{myblue}{rgb}{0,0.2,0.8}
\definecolor{mydarkgray}{rgb}{0.66,0.66,0.66}

\definecolor{freecolor}{rgb}{0.96,0.68,0.16}
\definecolor{frozoncolor}{rgb}{0.39,0.41,0.41}

\usepackage{hyperref}
\hypersetup{
    urlcolor=mydarkblue,
}
\definecolor{mydarkblue}{rgb}{0,0.08,0.8}

\usepackage[T1]{fontenc}
\usepackage[font=small,labelfont=bf,tableposition=top]{caption}

\DeclareCaptionLabelFormat{andtable}{#1~#2  \&  \tablename~\thetable}

\makeatletter
\newif\if@restonecol
\makeatother

\usepackage[ruled,vlined]{algorithm2e}
\usepackage{algpseudocode}

\def\equationautorefname~#1\null{Eq.~(#1)\null}
\newcommand{\aref}[1]{\hyperref[#1]{Appendix~\ref{#1}}}

\title{
Wearable-based Human Activity Recognition with Spatio-Temporal Spiking Neural Networks 
}

\author{%
  Yuhang Li, Ruokai Yin, Hyoungseob Park, Youngeun Kim, Priyadarshini Panda\\
  Department of Electrical Engineering, Yale University\\
  New Haven, CT 06511, USA\\
  {\small \texttt{\{yuhang.li, ruokai.yin, hyoungseob.park, youngeun.kim, priya.panda\}@yale.edu}}
}

\begin{document}

\maketitle

\begin{abstract}
We study the Human Activity Recognition (HAR) task, which predicts user daily activity based on time series data from wearable sensors. 
Recently, researchers use end-to-end Artificial Neural Networks (ANNs) to extract the features and perform classification in HAR. However, ANNs pose a huge computation burden on wearable devices and lack temporal feature extraction. 
In this work, we leverage Spiking Neural Networks (SNNs)—an architecture inspired by biological neurons—to HAR tasks. 
SNNs allow spatio-temporal extraction of features and enjoy low-power computation with binary spikes. 
We conduct extensive experiments on three HAR datasets with SNNs, demonstrating that SNNs are on par with ANNs in terms of accuracy while reducing up to 94\% energy consumption. 
The code is publicly available in \url{https://github.com/Intelligent-Computing-Lab-Yale/SNN_HAR}

\end{abstract}

\section{Introduction}

With the rapid development of smart devices such as phones and fitness trackers, sensing user activities or behavioral insights becomes more important for healthcare purposes. 
In this case, Human Activity Recognition (HAR) \cite{lara2012survey, vrigkas2015review, anguita2013public} seeks to predict the user activities using the smart devices' sensors such as accelerometer, gyroscope, electroencephalogram (EEG) sensor, etc. 
The objective of HAR includes sports injury detection,
well-being management, medical diagnosis, smart building solutions \cite{ramanujam2021human} and elderly care \cite{nweke2019data}.

Traditionally, researchers use hand-crafted features and simple classifiers for HAR tasks. Yet this type of method requires expert knowledge to get high-quality features. More recently, deep learning has been introduced to use end-to-end feature extraction, as well as classification~\cite{nweke2018deep}.  
They use convolutional layers in the Artificial Neural Networks (ANNs)~\cite{mnih2015human, ignatov2018real, wan2020deep} and optimize the model with gradient descent. 
However, ANNs use full precision (\ie 32-bit floating-point operations) computation and incur low sparsity, bringing huge computation complexity and energy consumption to wearable devices. 
In addition, ANNs use ReLU neurons that do not consider correlation in time. This choice may be sub-optimal, especially for time series data since it simply adapts the ANN regime from the image domain.

To overcome the above limitations, we utilize Spiking Neural Networks (SNNs)~\cite{tavanaei2019deep, roy2019nature, DENG2020294, panda2020toward, christensen20222022,kim2022neural,kim2022exploring,kim2020revisiting,kim2022privatesnn,deng2022temporal,li2022converting} combined with convolutional layers for dealing with time series data in HAR. 
The HAR can be benefited from SNNs in two aspects:
(1) SNNs take advantage of binary spikes~(either 0 or 1) and thus enjoy multiplication-free and highly sparse computation that lowers energy consumption on time-series data;
(2) SNNs can inherently model the temporal dynamics in time series data. The spiking neurons from SNNs maintain a variable called the membrane potential through time. As long as the membrane potential exceeds a pre-defined threshold, the neuron will fire a spike in the current time step. 
We verify our SNNs on three popular HAR datasets (UCI-HAR~\cite{anguita2013public}, UniMB SHAR~\cite{micucci2017unimib}, HHAR~\cite{stisen2015smart}) and compare them with ANNs baselines. 
Our SNNs can deliver the same or even higher accuracy than ANNs while reducing up to 94\% energy consumption.

The rest of this paper is organized as follows: \autoref{sec_method} describes the problem statement and our approach to spatio-temporal SNN. \autoref{sec_exp} provides the experimental results and visualization on three HAR datasets. 
Finally, we give our conclusion remark in \autoref{sec_conclude}.

\section{Method}
\label{sec_method}

\begin{figure}
    \centering
    \includegraphics[width=\linewidth]{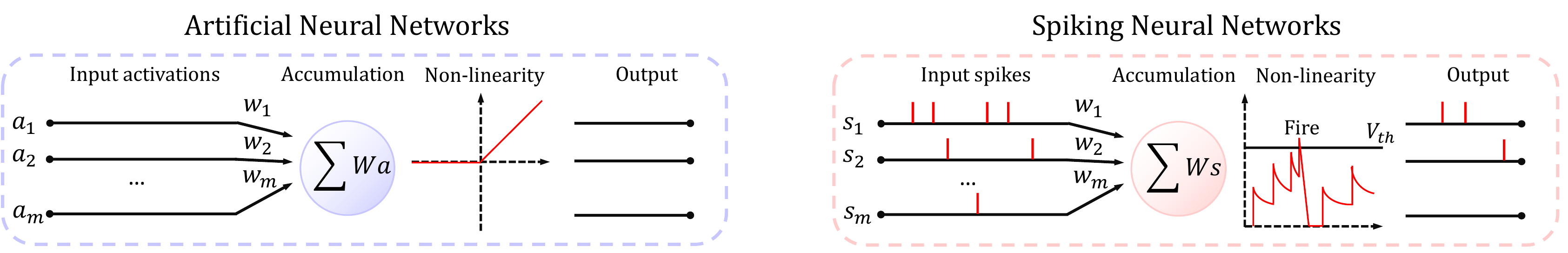}
    \caption{The schematic view of artificial neurons and spiking neurons. Artificial neuron takes full precision input and rectifies it if it is less than 0 and passes it otherwise; spiking neuron considers the correlation between times, and fires a spike only if the membrane potential is higher than a threshold.}
    \label{fig_neuron}
\end{figure}

\subsection{Notation and Problem Statement}
In this paper, vectors/matrices are denoted with bold italic/capital letters (\eg$\bm{x}\text{ and }\mathbf{W}$ represents the input vector and weight matrix). Constants are denoted by small upright letters. 

Concretely, we denote the wearable-based sensor dataset with $\{\mathbf{x}_i\}_{i=1}^N$, and each sample $\mathbf{x}_i\in\mathbb{R}^{T\times D}$ is collected when the wearer is doing certain activity $\mathbf{y}_i$, \eg running, sitting, lying, standing, \etc Here, data samples are streaming and have $T$ time steps in total. $D$ is the dimension of the sensor's output. As an example, the accelerometer records the acceleration in the $(x,y,z)$-axis, thus $D=3$ for the accelerometer data. 
We are interested in designing an end-to-end model $f(\cdot)$ and optimizing it to predict the accurate activity label $\mathbf{y}$.

\subsection{Spiking Neurons}

We adopt the well-known Leaky-Integrate-and-Fire (LIF) neurons model for spiking neurons \cite{liu2001spike}, which constantly receive inputs and outputs spikes through time.
Formally, the LIF neuron maintains the membrane potential $\bm{v}$ through time, and suppose at $t$-th time step ($1\le t\le T$), the membrane potential receives the pre-synaptic input charge $\bm{c}^{(t)}$, given by
\begin{equation}
    \bm{v}^{(t+1), \text{pre}} = \tau\bm{v}^{(t)} + \bm{c}^{(t)}, \text{where } \bm{c}^{(t)} = \mathbf{W} \bm{s}^{(t)}. \label{eq_lif}
\end{equation}
Here, $\tau$ is a constant between $[0, 1]$ representing the decay factor of the membrane potential as time flows, which controls the correlation between time steps. $\tau=0$ stands for 0 correlation and LIF degenerates to binary activation \cite{rastegari2016xnor} without temporal dynamics, while $\tau=1$ stands for maximum correlation and \cite{li2021free, deng2021optimal} proves that LIF will become ReLU neuron when $T$ is sufficiently large. 
$\bm{c}^{(t+1)}$ is the product between weights $\mathbf{W}$ and the spike $\bm{s}^{(t+1)}$ from last layer. 
After receiving the input charge, the LIF neuron will fire a spike if the pre-synaptic membrane potential exceeds some threshold, given by
\begin{equation}
\bm{s}^{(t+1)} = 
    \begin{cases}
        1 & \text{if } \bm{v}^{(t+1), \text{pre}} > V_{th} \\
        0 & \text{otherwise}
    \end{cases},
\label{eq_fire}
\end{equation}
where $V_{th}$ is the firing threshold. 
Note that the spike $\bm{s}^{(t+1)}$ will propagate to the next layer, here we omit the layer index for simplicity.

If the LIF neurons fire the spike, the membrane potential will be reset. This can be done by either soft-reset or hard-reset, denoted by 
\begin{equation}
    \begin{cases}
      \bm{v}^{(t+1)} = \bm{v}^{(t+1), \text{pre}}\cdot(1 - \bm{s}^{(t+1)})    & \texttt{\# Hard-Reset} \\
      \bm{v}^{(t+1)} = \bm{v}^{(t+1), \text{pre}} - \bm{s}^{(t+1)} \cdot V_{th} & \texttt{\# Soft-Reset}
    \end{cases},
\end{equation}
where hard-reset sets $\bm{v}^{(t+1)}$ to 0, while soft-reset subtracts $\bm{v}^{(t+1)}$ by $V_{th}$. We choose LIF neurons because the $s^{(t+1)}$ is binary and dependent on input in previous time steps. 
In our experiments, we will conduct ablation studies on the decay factor, the firing threshold, and the reset mechanism. 
In Fig.~\ref{fig_neuron}, we provide an overview difference between ANN and SNN neurons. 

\subsection{Integrating Spiking Neurons into Network}
We integrate spiking neurons into deep neural networks by replacing their non-linear activation with LIF. 
Specifically, since the time series data naturally has a time dimension, we also integrate the pre-synaptic potential charge along this time dimension. 
For instance, suppose $\bm{a}\in\mathbb{R}^{n\times c\times T}$ is a pre-activation tensor, where $n,c,T$ represent the batch size, channel number, and total time steps, respectively, we set the charge in each time step for LIF as the pre-activation in corresponding time step, \ie $\bm{c}^{(t)} = \bm{a}_{:,:,t}$.
Then, we stack the output spikes along the time dimension again, \ie $\mathbf{S}=\mathrm{stack}(\{\bm{s}^{(t)}\}_{t=1}^T)$, for calculating the pre-activation in next layer. 

\subsection{Optimization}
\newcommand{\FracPartial}[2]{\frac{\partial #1}{\partial #2}}

Although LIF neurons manage to model the temporal features and produce binary spikes, the firing function (\autoref{eq_fire}) is discrete and thus produces zero gradients almost everywhere, prohibiting gradient-based optimization. 
Particularly, the gradient of loss (denoted by $L)$ \wrt weights can be computed using the chain rule:
\begin{equation}
    \FracPartial{L}{\mathbf{W}} = \sum_{t=1}^T\FracPartial{L}{\bm{s}^{(t)}} \FracPartial{\bm{s}^{(t)}}{\bm{v}^{(t), \text{pre}}} \left(\FracPartial{\bm{v}^{(t), \text{pre}}}{\bm{c}^{(t)}} \FracPartial{\bm{c}^{(t)}}{\mathbf{W}}
    + \sum_{t^\prime=1}^{t-1}\FracPartial{\bm{v}^{(t), \text{pre}}}{\bm{v}^{(t)}}\FracPartial{\bm{v}^{(t)}}{\bm{v}^{(t^\prime), \text{pre}}}\FracPartial{\bm{v}^{(t^\prime), \text{pre}}}{\bm{c}^{(t^\prime)}} \FracPartial{\bm{c}^{(t^\prime)}}{\mathbf{W}}
    \right).
\end{equation}
Here, all other terms can be differentiated except $\FracPartial{\bm{s}^{(t)}}{\bm{v}^{(t), \text{pre}}}$ which brings zero-but-all gradients. To circumvent this problem, we use the surrogate gradient method \cite{li2021differentiable}. 
In detail, we use the triangle surrogate gradient, given by
\begin{equation}
\FracPartial{\bm{s}^{(t)}}{\bm{v}^{(t), \text{pre}}} =\max\left(0, 1 - \left|\frac{\bm{v}^{(t), \text{pre}}}{V_{th}} - 1\right|\right).
\end{equation} 
As a result, the SNNs can be optimized with stochastic gradient descent algorithms. 

\section{Experiments}
\label{sec_exp}
In this section, we verify the effectiveness and efficiency of our SNNs on three popular HAR benchmarks. 
We first briefly provide the implementation details of our experiments and then compare our method with ANNs' baselines.
Finally, we conduct ablation studies to validate our design choices.

\subsection{Implementation Details}

We implement our SNNs and existing ANNs with the PyTorch framework \cite{paszke2019pytorch}. For all our experiments, we use Adam optimizer \cite{kingma2014adam}. 
All models are trained for 60 epochs, with batch size 128. 
The only flexible hyper-parameter is the learning rate, which is selected from $\{1e-4, 3e-4, 1e-3\}$ with the best validation accuracy. We use Cosine Annealing Decay for the learning rate schedule. 
For all three HAR datasets, we split them to 64\% as the training set, 16\% as the validation set, and 20\% as the test set. We report test accuracy when the model reaches the best validation accuracy. Note that these datasets only have one label for each input sample, therefore top-1 accuracy is the same as the F-1 score. 
The dataset descriptions are shown below:

\textbf{UCI-HAR} \cite{anguita2013public} contains 10.3k instances collected from 30 subjects. It involves 6 different activities including walking, walking upstairs, walking downstairs, sitting, standing, and lying. The sensors are the 3-axis accelerometer and 3-axis gyroscope (both are 50Hz) from Samsung Galaxy SII. 

\textbf{UniMB SHAR} \cite{micucci2017unimib} contains 11.7k instances collected from 30 subjects. It involves 17 different activities including 9 kinds of daily living activities and 6 kinds of fall activities. The sensor is the 3-axis accelerometer (maximum 50Hz) from Samsung Galaxy Nexus I9250. 

\textbf{HHAR} \cite{stisen2015smart} contains 57k instances collected from 9 subjects. It involves 6 daily activities including biking, sitting, standing, walking, stair up, and stair down. The sensors are accelerometers from 8 smartphones and 4 smart watches (sampling rate from 50Hz to 200Hz).

\vspace{-0.5em}
\subsection{Comparison with ANNs}
\vspace{-0.5em}

\textbf{Task Performance. }For ANN baselines, we select CNN \cite{aviles2019coarse}, DeepConvLSTM \cite{mukherjee2020ensemconvnet}, LSTM \cite{wang2020human}, and Transformer~\cite{vaswani2017attention} architectures. 
For our SNNs, we integrate them into CNN and DeepConvLSTM. 
The architecture specifications can be found in our code. 
Each result is averaged from 5 runs (random seeds from 1000 to 1004) and includes a standard deviation value. 
We summarize the results in Table \ref{tab_acc}, from which we find the SNNs have higher accuracy than the ANNs. 
For example, on the UniMB SHAR dataset, SpikeCNN has a 1.7\% average accuracy improvement over its artificial CNN counterpart. 
Even more remarkably, the SpikeDeepConvLSTM (SpikeDCL) on the UCI-HAR dataset reaches 98.86\% accuracy, which is 1\% higher than DCL. Considering the accuracy is approaching 100\%, the 1\% improvement would be very significant.
For UCI-HAR and HHAR datasets, we find SpikeCNN has similar accuracy to CNN, instead, the SpikeDeepConvLSTM consistently outperforms DeepConvLSTM, indicating that SNNs can be more coherent with the LSTM layer. 
Regarding the standard deviation of accuracy, we find that SNNs are usually more stable than ANNs, except for only one case, SpikeCNN on UCI-HAR.

\begin{figure}[t]
 \centering
 \begin{subfigure}[b]{0.48\textwidth}
     \centering
     \includegraphics[width=\textwidth]{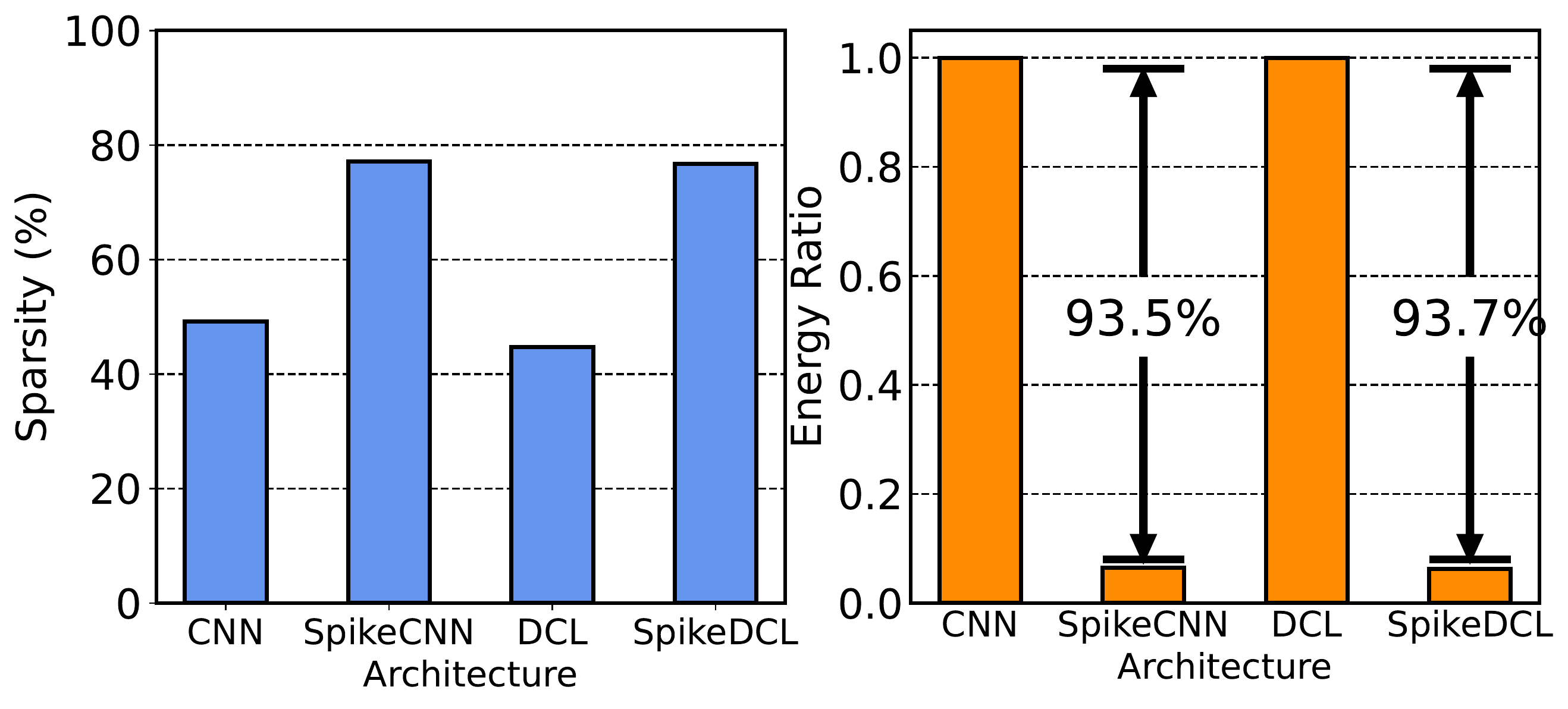}
     \caption{UCI-HAR}
     \label{fig:y equals x}
 \end{subfigure}
 \hfill
 \begin{subfigure}[b]{0.48\textwidth}
     \centering
     \includegraphics[width=\textwidth]{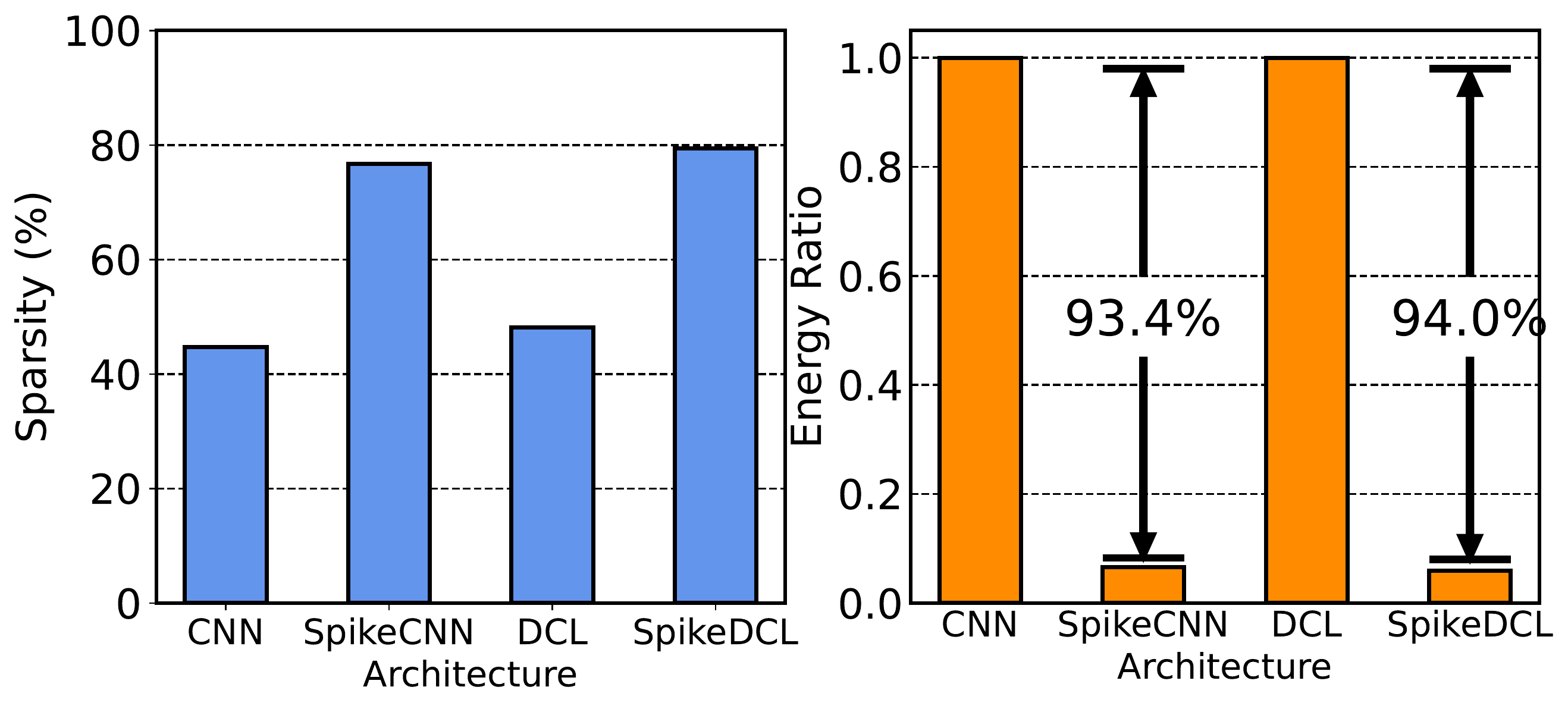}
     \caption{SHAR}
     \label{fig:three sin x}
 \end{subfigure}
 \hfill
    \caption{Hardware performance comparison between ANN and SNN. The ANN's energy consumption is normalized to 1. }
    \label{fig_hw}
    \vspace{-2em}
\end{figure}

\textbf{Hardware Performance.} 
Here, we compare two metrics, namely the activation sparsity and the energy consumption. 
Higher sparsity can avoid more computations with weights in hardware that supports sparse computation. 
We measure the sparsity either in ReLU (ANNs) or in LIF (SNNs) and visualize them in \autoref{fig_hw} left side.
The ReLU in ANN usually has around 50\% sparsity, an intuitive result since the mean of activation is usually around 0.
LIF neurons, however, exhibit a higher sparsity, approximately 80\%, probably due to the threshold for firing being larger than 0. 
As a result, the SNN has a higher potential to save more operations in inference.

The second metric in hardware performance is energy consumption. We estimate the energy consumption by evaluating the proposed SNN model together with our ReLU-based ANN baseline through the energy simulator proposed in \cite{yin2022sata}.
Particularly, we estimate the energy reduction ratio on the hardware accelerator \cite{yin2022sata}. 
The results are shown in \autoref{fig_hw} right side.
It can be seen that SNNs consume up to 94\% less energy than ANNs, which could largely promote the battery life in smart devices. 
In summary, SNNs bring higher task performance due to the LIF neurons, and also energy efficiency due to the  binary representation with high sparsity.

\newcommand{\mypm}[1]{\color{gray}{\small{$\pm$#1}}}

\begin{table}[t]
\centering
\caption{Accuracy comparison between different networks on three HAR datasets (DCL=DeepConvLSTM).}
\begin{tabular}{l | ccc ccc c | cc}
\toprule
{{Model}} &  CNN & DCL & LSTM & Transformer & SpikeCNN & SpikeDCL \\
\midrule
{UCI-HAR} \cite{anguita2013public} & 96.29\mypm{0.12} & 97.87\mypm{0.32} & 82.41\mypm{4.04} & 96.02\mypm{0.27} & \textbf{96.40\mypm{0.15}}  & \textbf{98.86\mypm{0.28}} \\
{SHAR} \cite{micucci2017unimib} & 92.38\mypm{0.51} & 90.78\mypm{1.05} & 83.87\mypm{0.96} & 83.19\mypm{0.74} &  \textbf{94.04\mypm{0.34}} & \textbf{92.08\mypm{0.77}} \\
{HHAR} \cite{stisen2015smart} & 96.19\mypm{0.14} & 97.15\mypm{0.17} & 95.59\mypm{0.20} & 95.82\mypm{0.16} & \textbf{96.20\mypm{0.09}} & \textbf{97.52\mypm{0.10}}\\ 
\bottomrule
\end{tabular}
\label{tab_acc}
\end{table}


\begin{table}[t]
\centering
\caption{Ablation study on the decay factor $\tau$.}
\begin{tabular}{l c p{2.5cm} c c c c c }
\toprule
\multirow{2}{*}{\bf Dataset} && \multirow{2}{*}{\bf Model} &  \multicolumn{5}{c}{\bf Decay Factor $\tau$}\\
\cmidrule(l{2pt}r{2pt}){4-8}
& & & 0.0 & 0.25 & 0.5 & 0.75 & 1.0 \\
\midrule
\multirow{2}{*}{UCI-HAR \cite{anguita2013public}} && SpikeCNN & 95.48 & 95.63 & 95.78 &\bf 96.40 & 95.92 \\ 
&& SpikeDCL & 94.36 & 96.50 & 97.57 & \bf 98.86 & 96.60 \\
\midrule
\multirow{2}{*}{SHAR \cite{micucci2017unimib}} && SpikeCNN & 93.54 & \bf 94.04 & 93.48 & 93.85 & 74.68 \\ 
&& SpikeDCL & 89.53 & \bf 92.08 &  90.93 &  90.10 & 60.55 \\
\bottomrule
\end{tabular}
\label{tab_decay}
\end{table}

\begin{table}[t]
\centering
\caption{Ablation study on the firing threshold $V_{th}$ and the reset mechanism.}
\begin{tabular}{l p{2.4cm} c c c c c c c}
\toprule
\multirow{2}{*}{\bf Dataset} &\multirow{2}{*}{\bf Model} &  \multicolumn{4}{c}{\bf Firing Threshold $V_{th}$} & \multicolumn{2}{c}{\bf Reset}\\
\cmidrule(l{2pt}r{2pt}){3-6} \cmidrule(l{2pt}r{2pt}){7-8} 
& & 0.25 & 0.5 & 0.75 & 1.0 & Hard & Soft\\
\midrule
\multirow{2}{*}{UCI-HAR \cite{anguita2013public}} & SpikeCNN & 95.71 & \bf 96.40 & 96.18 & 96.11 & 96.09 &\bf  96.40\\
& SpikeDCL & 98.27 & \bf 98.86 & 97.60 & 96.81 & 98.53 &\bf 98.73 \\
\midrule
\multirow{2}{*}{SHAR \cite{micucci2017unimib}} & SpikeCNN & 93.91 &\bf 94.04 & 93.89 & 93.87 & 92.75 & \bf 94.04 \\ 
& SpikeDCL & 91.42 & \bf 92.08 & 91.72 & 91.53 & 91.13 & \bf 92.08 \\
\bottomrule
\end{tabular}
\label{tab_fire}
\end{table}

\vspace{-0.5em}
\subsection{Ablation Studies}
\vspace{-0.5em}

In this section, we conduct ablation studies with respect to the (hyper)-parameters in the LIF neurons, including decay factor, threshold, and reset mechanism.
We test SpikeDCL and SpikeCNN on UCI-HAR and SHAR datasets. 

\textbf{Decay Factor.} 
We select 5 fixed decay factors from $\{0.0, 0.25, 0.5, 0.75, 1.0\}$. Note that as discussed before $\tau=0$ indicates no correlation between two consecutive time steps, therefore SNN becomes equivalent to Binary Activation Networks~(BAN), while $\tau=1$ indicates full correlation. 
We provide all results in \autoref{tab_decay}. We can find that $\tau$ has a huge impact on the final test accuracy.
For the UCI-HAR dataset with SpikeDCL, the accuracy of $\tau=0$ is 94.36\% while the accuracy of $\tau=0.75$ is 98.86\%.
Additionally, if we compare other $0<\tau<1$ cases with $\tau=0$, we find that $\tau=0$ always produces a large deficiency. 
\emph{This indicates that considering the temporal correlation with $\tau>0$ is necessary for the time series tasks. }
Moreover, for the SHAR dataset, the $\tau=1$ only has 60.55\% accuracy while the $\tau=0.25$ case achieves 91.72\% accuracy.

\textbf{Firing Threshold.}
We next study the effect of the firing threshold. Generally, the firing threshold is related to the easiness of firing a spike. 
We set the threshold in $\{0.25, 0.5, 0.75, 1.0\}$ and run the same experiments with the former ablation.
Here, through \autoref{tab_fire} we observe that the firing threshold has a unified pattern. SNN reaches its highest performance when the firing threshold is set to 0.5. 
This result is not surprising since 0.5 is in the mid of 0 and 1, and thus has the lowest error for the firing function (see Eq. (\ref{eq_fire})). 
Meanwhile, we find the difference in accuracy brought by the firing threshold is lower than the decay factor. 
For instance, the largest gap when changing the threshold for SpikeDCL on the SHAR dataset is 0.65\%, while this gap can be 32\% when changing the decay factor. 
Therefore, the SNN is more sensitive to the decay factor rather than the threshold. 

\textbf{Reset Mechanism.}
Finally, we verify the reset mechanism for SNNs, namely soft-reset and hard-reset. 
The results are sorted in the \autoref{tab_fire} as well. 
For all cases, the soft-reset mechanism is better than the hard-reset.
We think the reason behind this is that the hard reset will directly set the membrane potential to 0, therefore cutting off the correlation between two time steps. 
Instead, the soft reset keeps some information on membrane potential after firing.

\section{Conclusion}
\label{sec_conclude}

In this paper, we have introduced Spiking Neural Networks (SNNs) for HAR tasks, which, to our best knowledge, is the first of its kind study. 
Compared to the original Artificial Neural Networks~(ANNs), SNNs utilize their LIF neurons to generate spikes through time, bringing energy efficiency as well as temporally correlated non-linearity. Our results show that SNNs achieve competitive accuracy while reducing energy significantly.

\bibliography{ref}
\bibliographystyle{unsrt}

\end{document}